\documentclass[runningheads]{llncs}
\usepackage{graphicx}
\usepackage{color}
\usepackage{multirow}
\usepackage{booktabs}
\usepackage{amsmath}
\usepackage{bm}
\usepackage{hyperref} 
\usepackage{floatrow}
\usepackage{subfigure}
\usepackage{CJKutf8}
\usepackage{enumitem}
\newlist{numbers}{enumerate}{1}
\setlist[numbers, 1]{label = \emph{\arabic*)}}
\begin{document}

\title{Knowledge-aware Method for Confusing Charge Prediction}
\titlerunning{Knowledge-aware Method for Confusing Charge Prediction}
%
\author{Xiya Cheng\inst{1}
\and
Sheng Bi\inst{1}\thanks{Corresponding author.}
\and
Guilin Qi\inst{1}
\and
Yongzhen Wang\inst{2}
}

\authorrunning{Cheng et al.}
%
\institute{School of Computer Science and Engineering, Southeast University, Nanjing, China
\and
Indiana University Bloomington, Bloomington, IN, USA\\ 
\email{\{chengxiya, bisheng, gqi\}@seu.edu.cn, kuadmu@163.com}}

\maketitle              
\begin{abstract}
Automatic charge prediction task aims to determine the final charges based on fact descriptions of criminal cases, which is a vital application of legal assistant systems. Conventional works usually depend on fact descriptions to predict charges while ignoring the legal schematic knowledge, which makes it difficult to distinguish confusing charges. In this paper, we propose a knowledge-attentive neural network model, which introduces legal schematic knowledge about charges and exploit the knowledge hierarchical representation as the discriminative features to differentiate confusing charges. Our model takes the textual fact description as the input and learns fact representation through a graph convolutional network. A legal schematic knowledge transformer is utilized to generate crucial knowledge representations oriented to the legal schematic knowledge at both the schema and charge levels. We apply a knowledge matching network for effectively incorporating charge information into the fact to learn knowledge-aware fact representation. Finally, we use the knowledge-aware fact representation for charge prediction. We create two real-world datasets and experimental results show that our proposed model can outperform other state-of-the-art baselines on accuracy and F1 score, especially on dealing with confusing charges.

\end{abstract}
\section{Introduction}
Given a criminal case's fact description, automatic charge prediction task aims to teach a machine judge to identify the appropriate charge, such as \textit{theft}, \textit{robbery} or \textit{fraud}. It is a critical technique of intelligent legal judgment system. On the one hand, it can help human judges handle the heavy daily routine work and improve their work efficiency. On the other hand, people without legal background knowledge can consult the machine judge about legal guidance and assistance by describing a case they are worried about, which is low-cost but high-quality.

Automatic charge prediction has been investigated for many years, and most existing works treat charge prediction as a text classification problem. Early efforts~\cite{liu2004case,liu2006exploring} extract shallow textual features (e.g. characters, words, and phrases) artificially to predict charges, which are time-consuming and labor-intensive. Inspired by the big success of deep neural networks on natural language processing tasks, many methods based on deep learning have been explored for charge prediction. Luo et al.~\cite{luo2017learning} proposed an attention-based neural model for charge prediction by selecting the most relevant law articles to support charge problem. As for multi-label charge prediction, Wei et al.~\cite{wei2019external} proposed an external knowledge enhanced end-to-end multi-label charge prediction method with automatic label number learning and a number learning network.

However, there exist many confusing charges, such as $<$\textit{theft}, \textit{robbery}, \textit{defraud}$>$. The fact descriptions of these confusing charges only have some slight differences, which are difficult to capture. For instance, a \textit{robbery} case also contains the fact of \textit{theft}. The critical difference between these two charges is whether the defendant intended to harm the victim in his subjective will.
To solve this problem, some researchers began to consider introducing external information. Hu et al.~\cite{hu2018few} introduced several discriminative attributes of charges to alleviate the few-shot charges prediction. They constructed 10 discriminative legal attributes of charges as the internal mapping between fact descriptions and charges, which offer signals for distinguishing confusing charges. However, the drawback of their method is relying too much on experts and both summarizing and annotating attributes need lots of manual work.
Xu et al.~\cite{xu2020distinguish} proposed an end-to-end model to automatically learn subtle differences between confusing law articles and designed an attention mechanism that exploits the learned differences to attentively extract discriminative features from fact descriptions. Nevertheless, there is not sufficient information in the law articles to distinguish different charges, especially the confusing charges.
In practice, we would like to have a method that thinks like a legal domain expert and learns the elementary knowledge of how to distinguish different charges.

Therefore, we propose a novel knowledge-aware model to predict charges. In this model, we introduce the legal schematic knowledge about charges and exploit hierarchical knowledge representation as the discriminative features to differentiate confusing charges. These features can provide explicit knowledge about how to distinguish confusing charges. Specifically, our model takes the textual fact description as the input and learns the fact representation through Graph Convolutional Network (GCN). Meanwhile, we utilize the legal schematic knowledge transformer (LK-Transformer) to generate crucial knowledge representations oriented to the legal schematic knowledge at both the schema and charge levels. We apply a knowledge matching network for effectively incorporating charge information into the fact to learn knowledge-aware fact representation. Finally, we use the knowledge-aware fact representation for charge prediction. To validate the effectiveness of our model, we conduct a series of experiments on several real-world datasets. Comprehensive experiments show that our model outperforms other baselines and achieves significantly improvements for confusing charges. 

\section{Related Work}
The charge prediction task has drawn increasing attention in recent years. In early studies on charge prediction, most researchers inclined to formalize it as a text classification problem, which takes the fact description as input and outputs a charge label. Liu et al.~\cite{liu2004case,liu2006exploring} used K-Nearest Neighbor (KNN) and extracted shallow textual features (e.g. characters, words, and phrases) artificially to predict charges. By taking data from the European Court of Human Rights as an example, Medvedeva et al.~\cite{medvedeva2019using} addressed the potential in treating case law as quantitative data to predict judicial decisions and assessed how well Support Vector Machine (SVM) Linear Classifier is able to determine court judgments. However, these conventional methods can only leverage shallow textual features or manually designed factors, both need massive human efforts and hard to scale. 

Given the fact description, Luo et al.~\cite{luo2017learning} proposed a hierarchical attention network for charge prediction by selecting the most relevant law articles to predict charges. As for multi-label charge prediction, Wei et al.~\cite{wei2019external} proposed a knowledge enhanced end-to-end multi-label charge prediction method with automatic label number learning network. However, these works fail to answer why the prediction results are correct, and the prediction results are hard to interpret. Thus, Bruninghaus et al.~\cite{bruninghaus2003predicting} investigated an algorithm, IBP, which combines reasoning with an abstract domain model and case-based reasoning techniques to predict the outcome of case-based legal arguments. Further, SMILE+IBP~\cite{ashley2009automatically} is proposed to predict and explain the outcomes of case scenarios according to a database of previously classified cases. 
In recent years, Li et al.~\cite{li2019research} provided a cognitive computing framework for predicting judicial decisions, whose predicting results are interpretable in a way that induction rules are supplied. 

Nevertheless, all of these methods have poor performances on confusing charge prediction, which will influence the precise adjustment of prison term. Only few research works focus on confusing charges. Li et al.~\cite{li2019element} proposed an element-driven attentive neural network model, EDA-NN, which introduces the legal constitutive elements as the discriminative features to distinguish confusing charges. To improve prediction accuracy, ~\cite{yang2019legal} focused on word collocations information and integrated word collocations features of fact descriptions into the network via an attention mechanism. These works don't consider the charge information, which is proved to be useful in predicting charges~\cite{chen2019learning}. Fortunately, Hu et al.~\cite{hu2018few} constructed several discriminative attributes of charges as the internal mapping between fact descriptions and charges, which offer effective signals for distinguishing confusing charges. However, only a few confusing charges can be distinguished. It still remains a challenge to distinguish all confusing charges.

To address this issue, we propose a novel knowledge-aware framework to predict charges by incorporating legal schematic knowledge about charges. 

\section{Methodology}
\begin{figure}
    \includegraphics[width=0.88\textwidth]{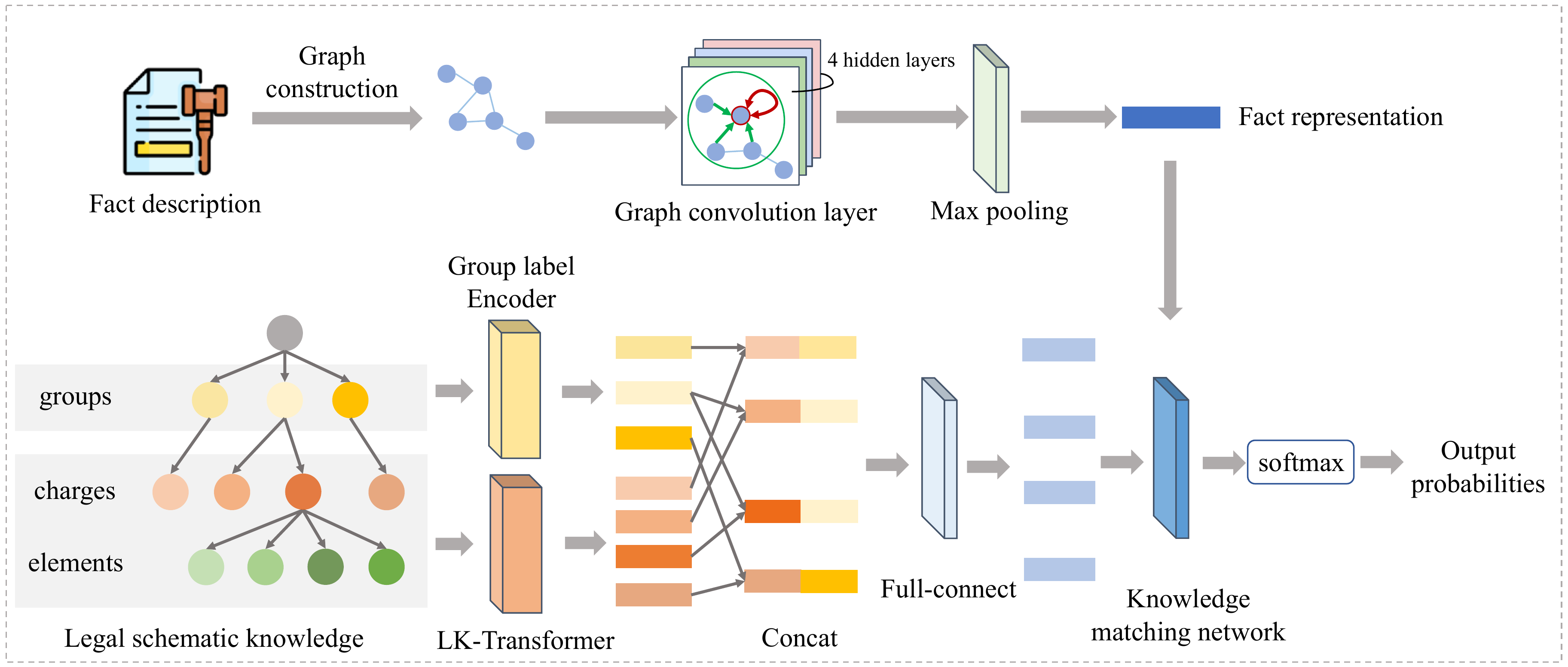}
    \caption{An illustration of our proposed model for charge prediction.} \label{fig1}
\end{figure}

In this section, we present the details of our model. The overall architecture of our model is shown in Fig.~\ref{fig1}. We input the distributed representation of a fact description to GCN to learn the fact representation. Then we utilize the LK-Transformer to generate crucial knowledge representations oriented to the legal schematic knowledge at both the schema and charge levels. To improve the uniqueness, the group label information is encoded and concatenated with corresponding knowledge representations. Moreover, we apply a knowledge matching network for effectively incorporating charge information into the fact to learn knowledge-aware fact representation. With the knowledge-aware fact representation, we use a softmax layer to output the predicted distributions of charges. 

In the following subsections, we first give the problem formulation and introduce our legal schematic knowledge. Then we describe the neural encoder of fact description and the knowledge-aware fact representation. In the end, we show the output layer and the loss function of our model.

\subsection{Problem Formulation}
For each case, the fact description is regarded as a word sequence $X=(x_1,x_2,...,x_n)$, where $n$ is the length of $X$. We extract legal schematic knowledge $K$ about $m$ charges from findlaw\footnote{http://china.findlaw.cn/zuiming/}. Given the fact description $X$ and legal schematic knowledge $K$, the charge prediction task aims to predict a charge label $y$.

\subsection{Legal Schematic Knowledge}
\begin{figure}
    \includegraphics[width=0.8\textwidth]{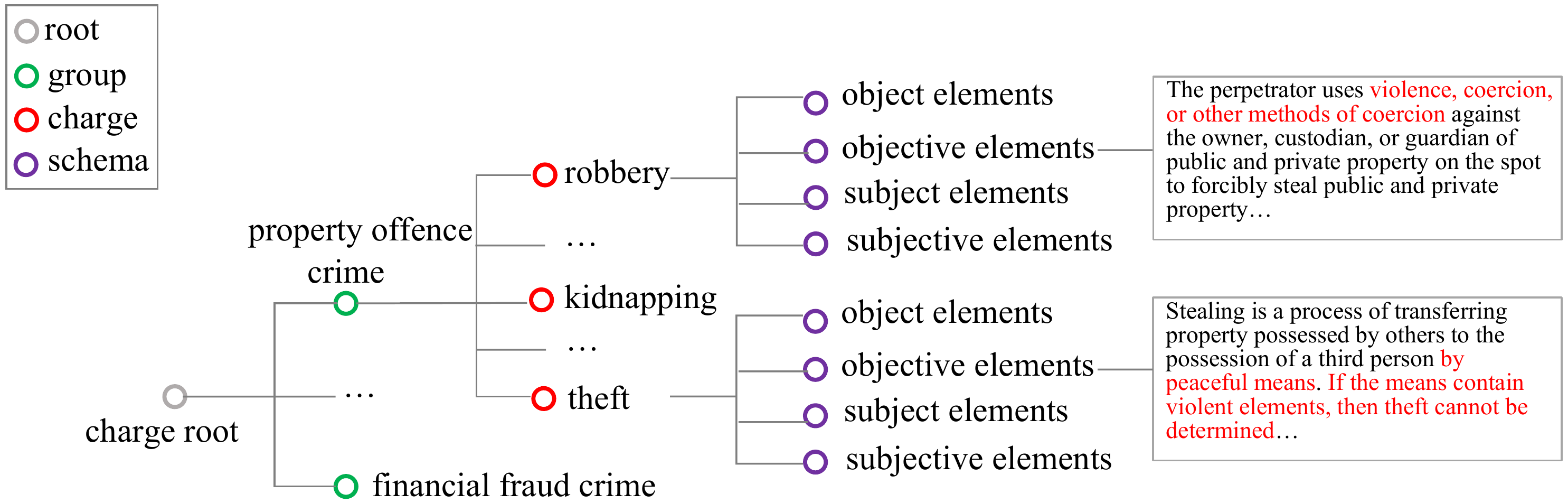}
    \caption{An illustration of the structure of legal schematic knowledge. Different charges can be distinguished by constitutive elements.} \label{fig2}
\end{figure}

The legal schematic knowledge, whose structure is presented in Fig.~\ref{fig2}, has four hierarchies. The root is charge object, whose children are charge groups. Groups are macro categories of all charges in general, such as ``crime of infringing property" and ``crime of financial fraud". Besides, the children of each group are specific charges, and many are confusing charges. Note that there are 25 groups and 435 charges in total. To distinguish confusing charges, we add the knowledge of constitutive elements to represent charges. Concretely, each charge has four schemas: object elements $K_1$, objective elements $K_2$, subject elements $K_3$ and subjective elements $K_4$. These schemas are the key points to define a charge and provide explicit knowledge about how to distinguish confusing charges.

\subsection{Fact Representation}
As illustrated in Fig.~\ref{fig1}, fact encoder encodes the discrete input sequence into hidden states, and obtain the fact representation by max pooling. Since traditional neural encoder, e.g., RNN and CNN, can only capture semantic and syntactic information in local consecutive word sequences well, while ignoring global word co-occurrence in a corpus which carries non-consecutive and long-distance semantics, we employ GCN to encode fact descriptions. GCN is a simple and effective graph neural network that can capture high order neighborhood information and rich global structure information. 

\label{sec3.2.1}
\subsubsection{Graph Construction} First, fact encoder converts each word $x_i \in X$ into its word embedding $x_i \in R^d$ through an embedding layer, where $d$ is the dimension of word embeddings. Then, we get the corresponding word embedding sequence as $W=(w_1,w_2,...,w_n)$, which can be used as the original feature matrix \textit{$H^{(0)}$}. After that, we build a graph for the given text. We regard all the words that appeared in the text as the nodes of the graph. The point-wise mutual information (PMI) is employed to calculate the weights of edges, which can preserve the global word co-occurrence information~\cite{yao2019graph}. Specially, we use a fixed-size sliding window on all documents in the source and legal domain to collect word co-occurrence statistics. 

The PMI of a word pair $w_i$, $w_j$ is computed as: 
    $p(w_i, w_j) =\frac{W(w_i, w_j)}{|W|}, \\
    PMI(w_i, w_j) =log\frac{p(w_i, w_j)}{p(w_i)p(w_j)}$, where $p(w_i)=\frac{W(w_i)}{|W|}$. $W(w_i)$ is the number of sliding windows that contain the word $w_i$. $W(w_i, w_j)$ is the number of sliding windows that contain both words $w_i$ and $w_j$, and $|W|$ is the total number of sliding windows. The PMI score can reflect the semantic correlation between words. If the PMI value is negative, there is little or no semantic correlation. Therefore, we only add edges between word pairs with positive PMI scores:
\begin{equation}
    a_{ij}=\left\{
    \begin{aligned}
    PMI(w_i, w_j) & , & PMI(w_i, w_j)>0, \\
    0 & , & PMI(w_i, w_j) \leq 0.
    \end{aligned}
    \right.
\end{equation}
where $a_{ij}$ is the relation between the word $w_{i}$ and $w_{j}$. After this process, we obtain the word relations $\mathcal{A}$ over the global corpus, and the adjacent matrix $A$ is the subset of $\mathcal{A}$ for each document.

\subsubsection{Graph Propagation} With graph representation, we can capture complex semantic and long-distance relation between words. Then we need update the representation of each node by message passing. The propagation rule of GCN can be interpreted as the Laplacian smoothing~\cite{li2018deeper}. The new feature of a node is calculated as the weighted average of itself and its neighbors’, followed by a linear transformation. Further, each node can collect and integrate messages from adjacent nodes to update its representation, which is defined as: $H^{(k+1)} = \phi(\widetilde{D}^{-\frac{1}{2}}\widetilde{A}\widetilde{D}^{-\frac{1}{2}}H^{(k)}W^{(k)})$, where $\widetilde{A} = A + I$ is the adjacency matrix $A$ with added self-connection $I$, $\widetilde{D}$ is a diagonal matrix with $\widetilde{D}_{ii} =\sum_{j}\widetilde{A}_{ij}$, \textit{$H^{(k)}$} and \textit{$W^{(k)}$} are the node representation matrix and the trainable parameter matrix for the $k$-th layer, \textit{$H^{(0)}$} is the original feature matrix, and $\phi(\cdot)$ is the activation function.

As shown in Fig.~\ref{fig1}, we use 4 hidden layers in GCN. After applying the propagation rule to our graph defined before, each node would learn a distributed representation in the last layer eventually. These representations are then input to a max-pooling layer to obtain the final fact representation $\vec{e}=[e_1, \cdots, e_s]$ as
    $e_i=\max(h_{1,i},\cdots,h_{n,i}), \forall i \in [1,s]$,
here, $s$ is the dimension of hidden states.

\subsection{Knowledge-aware Fact Representation}
To address confusing charges, we exploit the legal schematic knowledge representation about charges as the discriminative features to differentiate confusing charges. As we mentioned above, with legal schematic knowledge provided, we can capture some slight differences in specific factors of the criminal motive, action or consequence among confusing charges and further distinguish them.

\subsubsection{LK-Transformer}
\begin{figure}
\includegraphics[width=10cm]{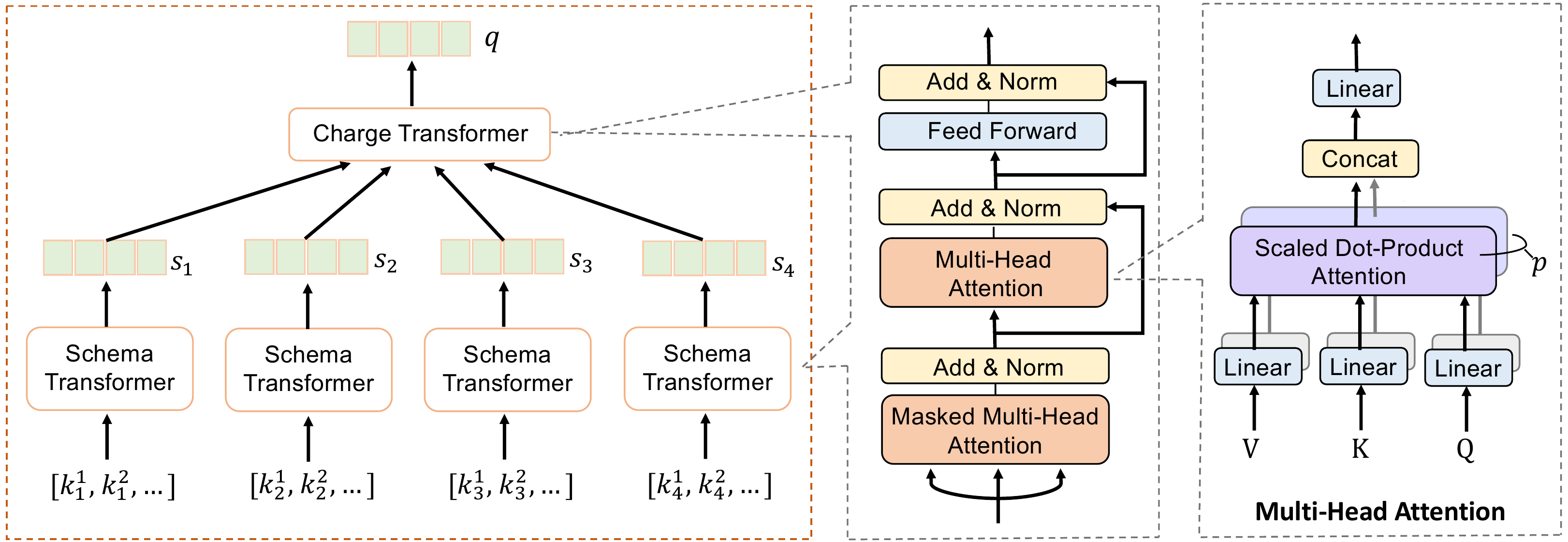}
\caption{An illustration of the LK-Transformer. The middle part is the explicit structure of transformer and the right part is the explanation of multi-head attention.} \label{fig3}
\end{figure}

Since the tree structure of our legal schematic knowledge, we utilize the LK-Transformer to generate crucial knowledge representations at both the schema and charge levels. The structure of our LK-Transformer is shown in Fig.~\ref{fig3}. For each charge, we input the schemas $K_1$, $K_2$, $K_3$ and $K_4$ to four schema transformers respectively and obtain corresponding semantic representations $s_1$, $s_2$, $s_3$ and $s_4$. Later, the four schema representations are fed into a charge-level transformer to compute the knowledge representation $q$. 

More specifically, the schema-level transformer and the charge-level transformer share the same model, shown in the middle halves of Fig.~\ref{fig3}. The first is a masked multi-head self-attention mechanism, the second is a multi-head self-attention mechanism, and the third is a position-wise fully connected feed-forward network. We employ a residual connection~\cite{he2016deep} around each of the three sub-layers, followed by layer normalization~\cite{ba2016layer}. That is, the output of each sub-layer is $LayerNorm(x + Sublayer(x))$, where $Sublayer(x)$ is the function implemented by the sub-layer itself.

 Attention mechanism softly maps a query and a set of key-value pairs to an output, where the query, keys, values, and output are all vectors. Taking queries and keys of dimension $d_k$, and values of dimension $d_v$ as input, it outputs a weighted sum of the values. As for the multi-head attention mechanism, multiple individual attention functions are performed in parallel, yielding $d_v$-dimensional output values. These are concatenated and once again projected, resulting in the final values, as depicted in the right halves of Fig.~\ref{fig3}. 
\begin{align}
    MultiHead(Q,K,V) &=Concat(head_1, \cdots, head_h)W^O, \\
    head_i &= Attention(QW_i^Q, KW_i^K, VW_i^V),
\end{align}
where the projections are parameter matrices $W_i^Q \in R^{d_{model} \times d_k}$, $W_i^K \in R^{d_{model} \times d_k}$, $W_i^V \in R^{d_{model} \times d_v}$ and $W^O \in R^{pd_v \times d_{model}}$. 

Besides, the fully connected feed-forward network (FFN) is consists of two linear transformations with a ReLU activation~\cite{nair2010rectified}.

\subsubsection{Knowledge–aware Fact Representation}
Through the LK-Transformer, we have obtained knowledge representations. To improve the uniqueness, we employ a \textbf{BiLSTM}\cite{zhang2015bidirectional} to encode the group label information. Then 
this representations are concatenated with corresponding knowledge representations to distinguish different charges and the final knowledge representations are $\bm{c}=(c_1,c_2,...,c_m)$.

\begin{figure}
    \includegraphics[width=7cm]{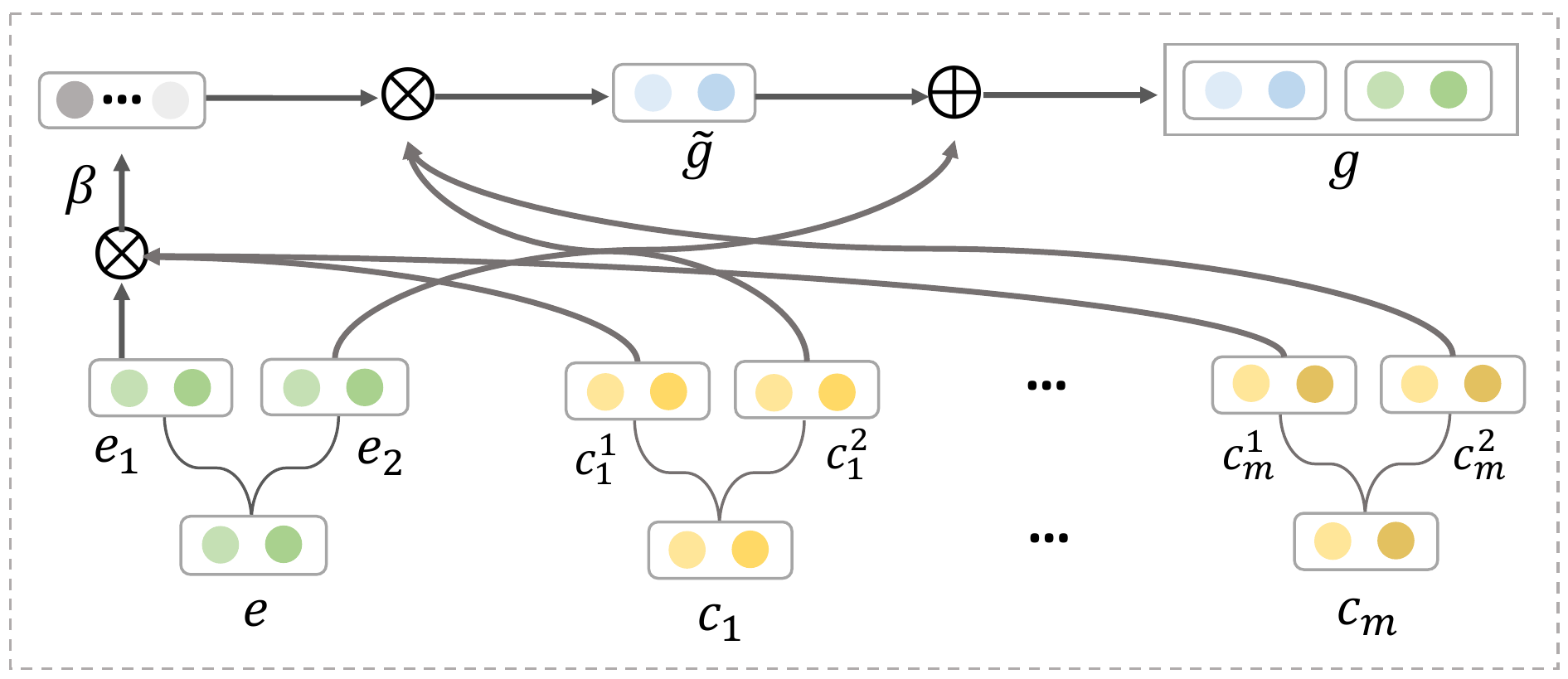}
    \caption{An illustration of the knowledge matching network. $\bm{e}$ denotes fact representation and $\bm{c_i}$ denotes each knowledge word representation, where $i \in [1,s]$. $\otimes$ means the multiply operation and $\oplus$ means the concatenate operation.} \label{fig4}
\end{figure}

We apply a knowledge matching network to select relevant information from knowledge and generate knowledge-aware fact representation. Let $\bm{e_j}=W^e_j\bm{e}$. Similarly, let $c_{i}^j=W^c_{ij}c_i$. As shown in Fig.~\ref{fig4}, the knowledge matching network first compute the attention $\bm{\beta}$ for the fact representation $\bm{e_1}$ and knowledge representation $\bm{c^1}$. Then the matched knowledge embeddings $\bm{\tilde{g}}$ for the fact are obtained by multiplying the attention matrix $\bm{\beta}$ with the knowledge representation $\bm{c^2}$. Finally, we concatenate the matched knowledge embeddings with the fact representation $\bm{e_2}$ to get the knowledge–aware fact representation $\bm{g}$.

Our knowledge matching function is defined as following:
\begin{equation}
    \bm{g}=\mathcal{F}(\bm{e_1},\bm{c^1},\bm{e_2},\bm{c^2})=Concat(\bm{e_2};\bm{\tilde{g}})=Concat(\bm{e_2};\bm{c^2}\bm{\beta}^T)
\end{equation}
where the function Concat is a concatenation operation, and $\beta$ is an attention score, computed by:
    $\bm{\beta} \propto exp(ReLU(W\bm{e_1})^TReLU(W\bm{c^1}))$,
where W is a weight matrix and ReLU is the rectified linear unit.

\subsection{Prediction and Optimization}
Ultimately, we use the knowledge-aware fact representation $\bm{g}$ to predict the final charge of a case in the output layer. The predicted probability distribution $y$ over all charges is calculated as $y =softmax(W^y \bm{g}+ b^y)$, where $W^y$ and $b^y$ are weight matrix and bias vector in the output layer. The training objective for our model is to minimize the cross-entropy between predicted charge $\widehat{y}$ and the ground-truth $y$. The charge prediction loss can be formalized as:
\begin{equation}
    Loss=-\sum_{i=1}^m y_i \cdot log(\widehat{y}_i).
\end{equation}

\section{Experiments}
In this section, we demonstrate the effectiveness of our model on criminal charges prediction. We conduct a series of experiments over two real-world datasets and compare our proposed model with several state-of-the-art baselines.

\subsection{Dataset Construction}
We collect 336,450 judgments of criminal cases published by the Chinese government from China Judgments Online\footnote{http://wenshu.court.gov.cn/}. Each case consists of several parts, such as the defendant’s information, fact description, charges. We only retain the fact description as input and the charge as label. The dataset is denoted as Criminal-All, which contains 203 different charges. Moreover, we select 80,000 cases which contain 86 confusing charges to form another dataset, named Criminal-Confusing. This dataset is designed for testing the ability of our model on predict confusing charges. For both Criminal-All and Criminal-Confusing, we randomly select 80\% of these cases for training, 10\% for validation, and 10\% for testing.

\subsection{Baselines}
To evaluate the performance of our framework, we compare our method with the following baselines:
\begin{itemize}
    \item[*] \textbf{HAN}: a Hierarchical Attention based RNN for document classification~\cite{yang2016hierarchical}.
    \item[*] \textbf{DPCNN}: a low-complexity word-level deep convolutional neural network architecture for text categorization~\cite{johnson2017deep}.
    \item[*] \textbf{Few-shot}: an attribute-based multi-task learning model for charge prediction by introducing discriminative legal attributes into consideration~\cite{hu2018few}.
    \item[*] \textbf{EDA-NN}: an element-driven attentive neural network model which can jointly predict the legal constitutive elements and judgment results~\cite{li2019element}.
    \item[*] \textbf{MPBN}: a multi-perspective bi-Feedback network with the word collocation attention mechanism for confusing charge prediction~\cite{yang2019legal}. 
    \item[*] \textbf{LADAN}: a Law Article Distillation based Attention Network to distinguish confusing charges~\cite{xu2020distinguish}.  
    \item[*] \textbf{Ours-GCN}: our model which replaces GCN with a Bi-LSTM layer.
    \item[*] \textbf{Ours-LK}: our model without legal schematic knowledge.
\end{itemize}

\subsection{Experimental Settings}
We first employ jieba\footnote{https://github.com/fxsjy/jieba} for Chinese word segmentation and word2vec~\cite{mikolov2013efficient} to train the word embeddings on all the legal judgments. We set all the hidden state size to 300. For the training, we use Adam~\cite{kingma2014adam} as the optimizer and set the learning rate to 0.001. We set the batch size and dropout rate to 128, 0.5, respectively. We repeat the training iterations until the difference between two continuous iterations is small enough. For the evaluation, we employ accuracy (Acc.), macro-recall (MR), macro-F1 (Ma-F1) and micro-F1 (Mi-F1) as metrics.

\subsection{Results and Discussion}
\begin{table*}[]
    \scriptsize
    \centering
    \caption{Charge prediction results of two datasets.}
    \label{tab1}
    \resizebox{!}{19mm}{ 
    \renewcommand\tabcolsep{4.0pt} 
    \begin{tabular}{lcccccccc}
    \toprule
    Datasets & \multicolumn{4}{c}{Criminal-All} & \multicolumn{4}{c}{Criminal-Confusing}  \\
     \cmidrule(r){2-5} \cmidrule(r){6-9} 
    Metrics
    &  Acc.      & MR     &   Ma-F1  &    Mi-F1
    &  Acc.      & MR     &   Ma-F1  &    Mi-F1  \\
    \midrule
    HAN & 0.858 &0.710 & 0.727 & 0.673 & 0.761 &0.612 & 0.631 & 0.582\\
    DPCNN & 0.878  &0.767 & 0.783 &0.719 & 0.782  &0.653 & 0.675 &0.614 \\
    Few-shot & 0.891 &0.824 & 0.842 & 0.788  & 0.844 &0.783 & 0.798 & 0.762\\
    EDA-NN & 0.901  &0.843 & 0.866 & 0.803 & 0.863  &0.801 & 0.814 & 0.787 \\
    MPBN & 0.903  &0.858 & 0.875 & 0.810 & 0.870  &0.811 & 0.826 & 0.792 \\
    LADAN & 0.905  &0.862 & 0.879 & 0.815 & 0.875  &0.817 & 0.829 & 0.797 \\
    \midrule
    Ours-GCN & 0.910  &0.877 & 0.889 & 0.822  & 0.882  &0.826 & 0.835 & 0.801 \\
    Ours-LK & 0.902 &0.843 & 0.862 & 0.799 & 0.864 &0.792 & 0.807 & 0.778 \\
    \textbf{Ours}& \textbf{0.914} & \textbf{0.893} &\textbf{0.909} & \textbf{0.834} & \textbf{0.898} & \textbf{0.841} &\textbf{0.856} & \textbf{0.819}\\
    \bottomrule
    \end{tabular}
    }
\end{table*}

As shown in Table~\ref{tab1}, we can observe that our model performs the best and significantly outperforms all baseline models on Criminal-All, which showcases the robustness and effectiveness of our proposed method for charge prediction. 

Our method is characterized by the incorporation of GCN and legal schematic knowledge. To verify the effectiveness of these modules, we design an ablation test respectively to investigate the effectiveness of these modules. As shown in Table~\ref{tab1}, we can observe that the performance drops obviously after removing the GCN or legal schematic knowledge. The Ma-F1 score decreases at least 2\%. Therefore, we conclude that both GCN and legal schematic knowledge play irreplaceable roles in our model and a combination of them will achieve better results on the charge prediction task.

To further investigate the effectiveness of our model on handling confusing charges, we show the performance on a confusing charge dataset, Criminal-Confusing, in Table~\ref{tab1}. We compare our model with the LADAN, which is the state-of-art model on predicting confusing charges. The results show that our model enhances the performance by about 2.3\%, 2.4\%, and 2.7\% relatively on Acc., MR, Ma-F1, respectively, which demonstrates the ability of our model.

\subsection{Case Study}
\begin{figure}
    \includegraphics[width=0.75\textwidth]{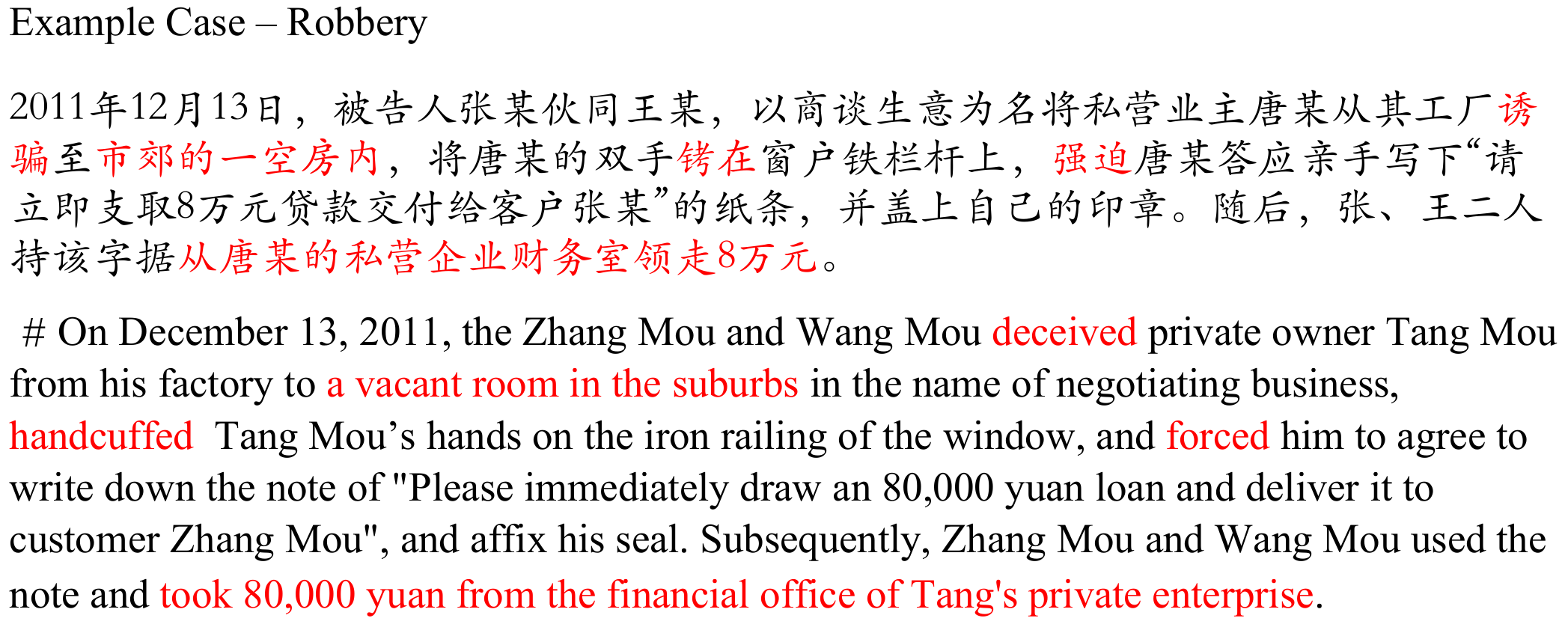}
    \caption{An illustration of the example case, whose golden charge is robbery.} \label{fig5}
\end{figure}

As shown in Fig.~\ref{fig5}, we choose a representative case to give an intuitive illustration of how the legal schematic knowledge improves the performance of confusing charge prediction. In the case, the defendant is convicted of robbery. It is often difficult to decide whether to judge a case as \textit{robbery} or \textit{kidnapping} since they are both related to violence and illegal possession. According to the constitutive requirements, one important difference between them is that the defendant intends to rob property directly from the victim in \textit{robbery}, while the defendant can only ask for property from a third party other than the victim in \textit{kidnapping}.

Therefore, the constitutive requirements are essential in the confusing charge prediction. From Table.~\ref{tab2}, we can see that only $Ours$ correctly predicts the charge of the case as \textit{robbery}. In contrary, other models predict it as \textit{kidnapping} incorrectly. Taking the Few-shot model as an example, although this case includes several attributes, like profit purpose, violence, public place and illegal possession, these attributes are not detailed enough to distinguish the charge. However, thanks to the legal schematic knowledge, our model can capture the key patterns and semantics relevant to constitutive requirements, which are assigned red color in Fig.~\ref{fig3}. Then the prediction charge is determined by these key patterns and semantics. This further demonstrates that our model is good at dealing with confusing charges through legal schematic knowledge. 
\begin{table*}[]
    \scriptsize
    \centering
    \renewcommand\tabcolsep{2.0pt} 
    \caption{Charge prediction results of the example case.}
    \label{tab2}
   
    \begin{tabular}{lcccccccc}
    \toprule
    Models & Gold & HAN & DPCNN & Few-shot & EDA-NN  & MPBN &LADAN & \textbf{Ours} \\
    \midrule
    case & \color{red}{\textit{robbery}}  & kidnapping & kidnapping & kidnapping & kidnapping  & kidnapping & kidnapping &\color{red}{\textit{robbery}} \\
    \bottomrule
    \end{tabular}
    
\end{table*}

\section{Conclusion}
In this paper, we focus on the task of charge prediction for criminal cases with confusing charges. To the best of our knowledge, we are the first to utilize legal schematic knowledge of charges as discriminative features to predict charges. We propose a legal schematic knowledge-aware framework to solve confusing charges. Our model adopts GCN and  LK-Transformer to learn better representations from informative case fact descriptions and legal schematic knowledge, respectively. Moreover, we utilize a knowledge matching network to learn knowledge-aware fact representation to improve the accuracy of confusing charge prediction. The experimental results on real-world datasets show that our model achieves significant improvements over baselines on all evaluation metrics for criminal cases, especially with confusing charges.

\section*{Acknowledgement}
Research presented in this paper was partially supported by the National Key Research and Development Program of China under grants (2018YFC0830200, 2017YFB1002801), the Natural Science Foundation of China grants (U1736204), the Judicial Big Data Research Centre, School of Law at Southeast University.

%
%
%
\bibliographystyle{splncs04}
\bibliography{nlpcc.bib}

\end{document}